# FinGAN: Generative Adversarial Network for Analytical Customer Relationship Management in Banking and Insurance

Prateek Kate[1,2], Vadlamani Ravi[1*] and Akhilesh Gangwar[3]

[1]Center of Excellence in Analytics,

Institute for Development and Research in Banking Technology

Castle Hills Road 1, Masab Tank, Hyderabad-500057 India.

[2]School of Computer and Information Sciences, University of Hyderabad-500046, India

[3]Mathworks, Hyderabad, India

prateek.kate111@gmail.com ; rav_padma@yahoo.com;  gangwar.akhilesh1993@gmail.com

**Abstract**: Churn prediction in credit cards, fraud detection in insurance, and loan default prediction are important analytical customer relationship management (ACRM) problems. Since frauds, churns and defaults happen less frequently, the datasets for these problems turn out to be naturally highly unbalanced. Consequently, all supervised machine learning classifiers tend to yield substantial false-positive rates when trained on such unbalanced datasets. We propose two ways of data balancing. In the first, we propose an oversampling method to generate synthetic samples of minority class using Generative Adversarial Network (GAN). We employ Vanilla GAN [1], Wasserstein GAN [2] and CTGAN [3] separately to oversample the minority class samples. In order to assess the efficacy of our proposed approach, we use a host of machine learning classifiers, including Random Forest, Decision Tree, support vector machine (SVM), and Logistic Regression on the data balanced by GANs. In the second method, we introduce a hybrid method to handle data imbalance. In this second way, we utilize the power of under- sampling and over-sampling together by augmenting the synthetic minority class data oversampled by GAN with the undersampled majority class data obtained by one-class support vigor machine (OCSVM) [4]. We combine both over-sampled data generated by GAN and the data under-sampled by OCSVM [4] and pass the resultant data to classifiers. When we compared our results to those of Farquad et al. [5], Sundarkumar, Ravi, and Siddeshwar [6], our proposed methods outperform the previous results in terms of the area under the ROC curve (AUC) on all datasets.

---

# 1. Introduction

Customer relationship management (CRM) is a process of analyzing and utilizing marketing databases and leveraging communication technologies for analyzing the corporate practices and methods to maximize the lifetime value of all customers [39]. Operationalization of CRM involves solving a host of business problems analytically. Some of the important CRM-related business problem include fraud detection, default prediction, churn prediction, and network intrusion detection. Churn is a scenario where a few of a bank's/financial institution's/service provider's currently loyal customers attrite or depart to competition. The number of banks and financial institutions has risen in the previous 2 decades. Customers shifting their allegiance from one bank to another has become a frequent occurrence. As a result, creating a robust and adaptable system that can anticipate/predict which existing loyal customers may churn out in the near future is very critical to improving the bottom line. Insurance fraud inflicts massive losses on the insurance firms to the tune of billions of dollars. Fraudulent claims are enormously expensive as well as entail reputational loss to insurance companies. Hence, detecting them, whether online or offline, will substantially minimize the loss. But since insurance firms provide relatively little data about fraud, identifying fraud is difficult. Then, loan defaults also cause a major dent to the revenues and operations of a bank. If only, loan defaults or non-performing assets (NPAs) are identified accurately, it will be beneficial to the bank both financial and reputational aspects.

Since the churn prediction, insurance fraud detection and default prediction datasets are much skewed, by virtue of the nature of the business, the distribution of positive class namely, churn/fraud/default versus the negative class namely, loyal/genuine/ regular classes follows a 90%:10% ratio or greater. This aspect is known as data imbalance in data science parlance. In binary classification, almost all machine learning classifiers underperform on such an unbalanced training set. That means, the algorithms become biased towards majority class such that the minority class samples are incorrectly predicted as the majority class primarily because of the overwhelmingly disproportionate large number of negative class samples vis-à-vis that of the positive class, which is of interest to us. It is hard to create synthetic data which has the same distribution as the original data.

In the literature, numerous approaches were proposed to oversample minority class data. The drawback with these strategies is that they repeat the data without understanding the distribution of class labels. Since GANs learn the distributional properties of any data and can generate fake samples, they are a reasonable candidate for generating fake samples from fraud class, churn class and default class. GANs are complex to train and deploy in production because they require a large amount of hyperparameter tweaking and distributed training assistance. We forecast fraud in auto insurance, consumer attrition in credit cards, and loan default customers by first proposing an oversampling of minority classes employing the deep learning architecture GAN [1] to deal with the problem of data imbalance, and then invoke a few machine learning classifiers for classification. Secondly, we propose another hybrid method in which we oversample the minority class samples using GAN and undersample the majority class by extracting the support vectors from majority class samples using one class support vector machine (OCSVM) to form a balanced dataset.

## 2. Literature Review

Nowadays, generative deep learning models have been widely used for synthetic data generation and after massive success of GAN [1] on image data it has attracted researchers to apply it to generate minority tabular data as well. Gangwar and Ravi [7] employed a variety of methods to oversample the minority class to address the data imbalance issue in the case of credit card fraud detection. They employed a couple of variants of GAN (vanilla GAN and Wasserstein GAN (WGAN) ) on the credit card fraud dataset. They also performed an ablation study using GAN+SMOTE, WGAN+SMOTE, GAN+ADASYN, WGAN+ADASYN, and validated using a couple of machine learning algorithms. The focus of their study is to reduce the false positives. Sisodia et al. [8] analyzed the performance of class balancing approaches for detecting credit card fraud. They used different SMOTE versions and performed undersampling & oversampling and validated their methods using different classifiers. Randhwa et al. [9] employed a diverse range of machine learning classifiers and designed an ensemble using majority voting for credit card fraud detection. Tanaka et al. [10] used GAN for oversampling the minority class data in the medical domain. They concluded that it is effective in the context of sensitive data privacy management. Mottini et al. [11] proposed Cramer GAN for PNR Synthesis of data consisting of categorical and numerical data. Instead of sigmoid they applied softmax to generate categorical variables. Fiore et al. [12] used Vanilla GAN to improve the efficacy of the detection of credit card fraud. Vega-Marquez et al. [13] deployed the conditional GAN to generate synthetic data for both classes. They discovered that the synthetic data improved F1-score with the xgboost classifier.

Che and Li [14] designed a new GAN named MaliGAN to generate discrete variables. By creating unique functions, they aim to create a differential model. Kusner and Hernández-Lobato [15] used gumbel-softmax distribution to create the discrete element sequence. Ping et al. [16] introduced a web-based data synthesizer using a Bayesian network to model the correlation between features. Various GAN models handling the tabular data have recently appeared in literature. RGAN and RCGAN [17] can generate real-valued time-series data. Camino et. al. [18] used GAN to generate multi-categorical samples. Choi et al. [19] used GAN for generating multi-label discrete patient records which combines an auto-encoder and a GAN to generate heterogeneous non-time-series continuous and/or binary data. This is also known as MedGAN. Patel et al. [20] designed a new GAN called CorGAN which learns the distribution of real patient records using convolutional generative adversarial networks. They demonstrated that CNNs are better than MLPs in extracting correlated features. Park et al. [21] proposed a novel architecture for GAN that employs CNN to produce tabular data to address the problem of generating synthetic data for a tabular dataset. By lowering cross-entropy loss, TableGAN explicitly enhances prediction accuracy on synthetic data. Xu and Veeramachaneni [22] proposed a GAN for generating tabular data which learns the marginal distribution of each column by minimizing the KL divergence. Xu et al. [3] proposed a new GAN for modeling synthetic data using conditional GAN known as CTGAN. CTGAN [3] aims to address the problem of several modes in continuous columns and unbalanced modes in discrete columns, which makes modeling complex.

Regarding churn prediction, Smith and Gupta [23] used MLP, Hopfield neural network, and self-organizing neural networks to address churn issues. Ferreira et al. [24] used a wireless data form Brazil, was evaluated using a multilayer perceptron, C4.5 decision tree, hierarchical neuro-fuzzy system as well as a data-mining tool called *rule evolver*, which is genetic algorithm- (GA) based. Kumar & Ravi [25] used data mining to conduct tests on credit card churn prediction at a bank. They employed balancing techniques such as SMOTE [26], undersampling, oversampling, and combination of both the methods prior to calling multilayer perceptron, logistic regression, random forest, radial basis function network and support vector machine. Larivie're & Van den Poel [27] found that there are two 'vital' churn periods in the financial services industry: the initial few years since becoming a member and the second couple of years after being a client for a lot longer. Ali [28] introduced a dynamic churn prediction framework. Vrebeke [29] proposed C4.5 and Ripper. Other studies include hybrid of backpropagation neural network (ANN) and self-organizing map (SOM) [30]. Sundarkumar and Ravi [31] presented a new approach that employed k Reverse Nearest Neighbor and one-class support vector machines (OCSVM) in tandem to tackle the data imbalance problem while also eliminating outliers. To test their suggested method, they employed the Insurance Fraud and Credit Card Churn datasets. Sundarkumar et. al. [6] proposed OCSVM based undersampling on the dataset used in [31]. Farquad et al. [5] proposed SVM-based active learning for undersampling and Naive Bayes Tree. Phua et al. [32] presented an approach for detecting fraud, which involved stacking to select the optimal base classifier and then to enhance outputs, integrated the predictions of these base classifiers. Stacking-bagging was the name given to this practice. Sublej et al. [33] presented a network algorithm on graphs for detecting automobile insurance frauds. However, a majority of the aforementioned research did not acknowledge the class imbalance concerns.

## 3 Proposed Methodology

First, we examine the full dataset and use data preparation techniques such as feature standardization. We split the data using stratified random sampling and the hold-out method in the ratio of 80%:20%. During the training, we did not touch the test data since they represent the reality.

### 3.1 Architecture of GAN

The main constituents of GAN namely, the generator and the discriminator are feedforward networks. As our data is structured we used multi-layer perceptron (MLP) for both the generator and discriminator. We also used the WGAN [2] and CTGAN [3] for the structured data generation. The sole difference between the vanilla GAN and the WGAN is the objective function, Therefore, we utilized the following architectures for the generator and discriminator. CTGAN [3] was used to produce synthetic data with no changes to its architecture. The architecture of vanilla GAN is depicted in Fig 1.

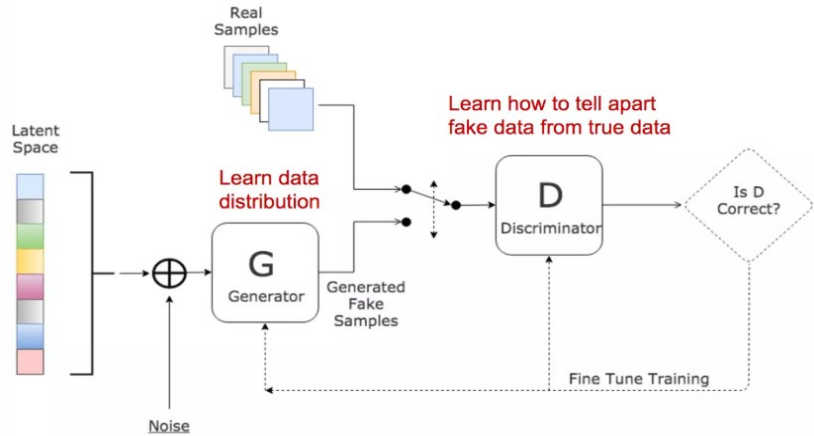

Fig.1 Architecture of GAN

### 3.3.1 Architecture of the Generator

**Input Layer:** This layer receives latent input that is produced at random and has a fixed length. We feed the generator with random noise, causing it to generate new synthetic samples each time.

**Hidden Layer:** Following approach has been used in the generator. We generate the synthetic data in two branches. First branch is used to generate the categorical variables, using softmax activation function. Because softmax produced categorical characteristics as a probability distribution across all potential labels. Second branch is used to generate the numerical variable, using sigmoid activation function.

**Output Layer:** Output layer used a concatenation function to concatenate the output generated by the two branches of the generator.

### 3.3.2 Architecture of the Discriminator

**Input Layer:** The discriminator's input layer receives a vector with the same length as the original data.

**Hidden Layers:**

1. First hidden layer has 128 neurons with leaky ReLU activation.
2. Second hidden layer has 64 neurons with leaky ReLU activation.
3. Third hidden layer has 32 neurons with leaky ReLU activation.
4. Fourth hidden layer has 16 neurons with leaky ReLU activation.
5. Fifth hidden layer has 8 neurons with leaky ReLU activation.

**Output Layer:** A binary output is produced which has a sigmoid activation function.

## 3.2. GAN + Classifiers

The schematic of our proposed model is depicted in Fig 2. We employed different architectures of GAN such as Vanilla GAN [1], WGAN [2], and CTGAN [3] for generating the synthetic records of the minority classes in order to balance the datasets. After generating the synthetic data, we merge the data oversampled data with the majority class data and the original minority class samples. Then, we feed this balanced data to the different machine learning classifiers such as RF, LR, etc. to check the effectiveness of our proposed approach.

## 3.3 GAN + OCSVM + Classifiers

The schematic of our proposed architecture is depicted in Fig 3. We proposed a novel hybrid approach in which we extract the support vectors from the majority class using OCSVM [4], which amounts to undersampling the majority class and perform oversampling of the minority class data using GAN. We chose OCSVM because it classifies the majority class, which entails the support vectors. If we choose the support vectors, then it amounts to performing row dimension reduction of the dataset . As a result, we achieve implicit undersampling. These support vectors are then combined with the GAN-oversampled minority class and the original minority class samples to produce a balanced dataset. To classify the balanced dataset and test the efficacy of our proposed model, we employed a variety of machine learning methods.

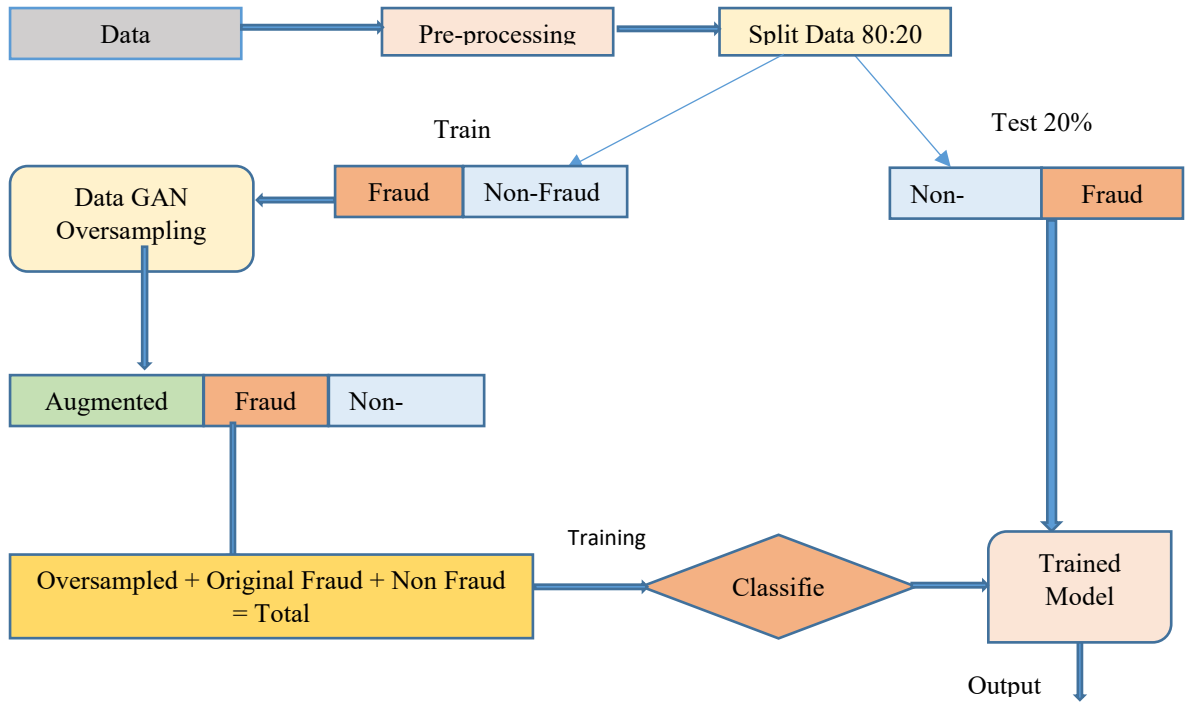

Fig 2. Schematic diagram of Model GAN + Classifier

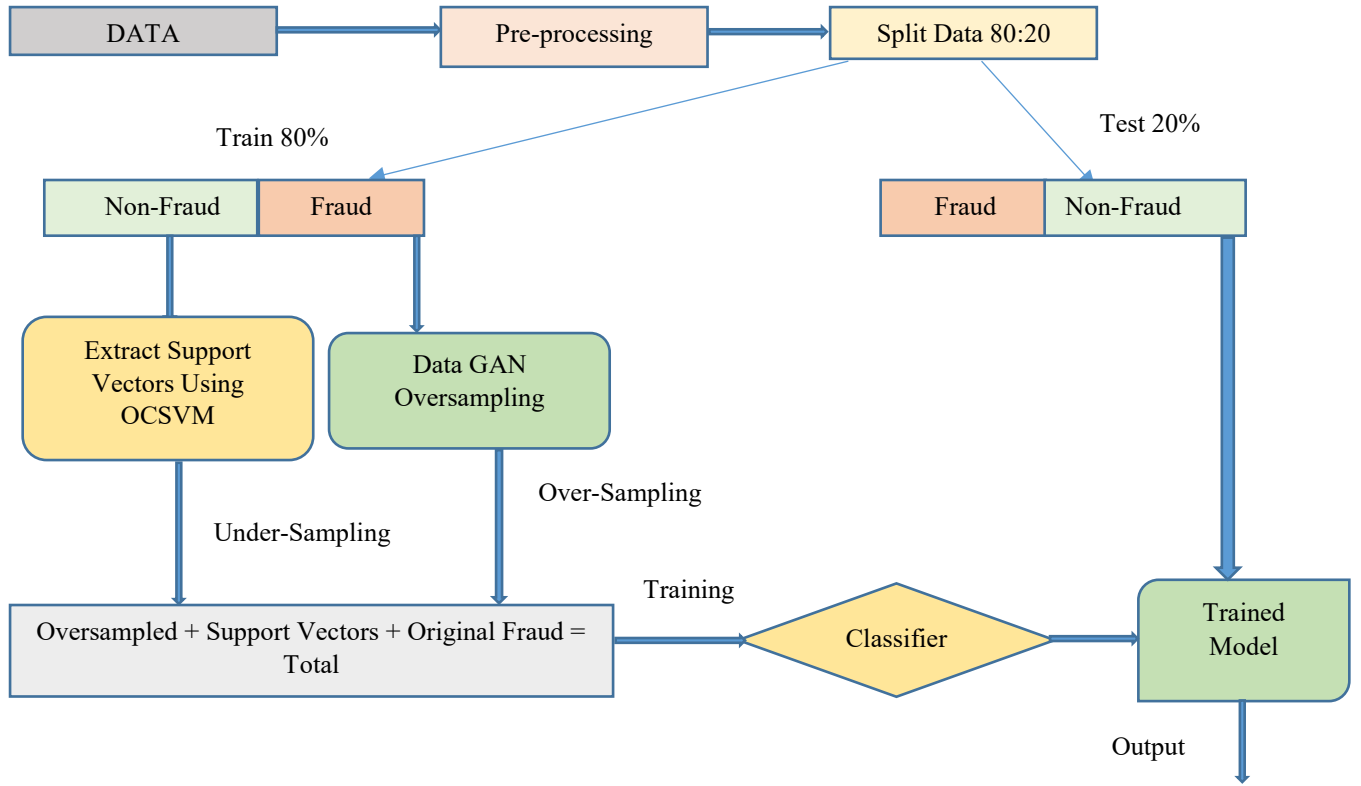

Fig 3. Schematic diagram of Model GAN + OCSVM + Classifier

### 3.4 Classifiers employed

In this study, we chose a variety of classifiers such as logistic regression (LR), support vector machines (SVM), decision trees (DT), Random Forest (RF), XGBoost, LightGBM, and Multilayer perceptron (MLP) for final classification prupose. Random forest and XGBoost are ensemble techniques used for classification and regression problems. In ensemble techniques we create multiple models and combine the results of the models to produce the results. In the case of classification, it uses majority vote to get the results whereas for the regression it uses statistical measures such as mean and median. The trained classifiers are tested on the test data.

### 3.5. Experimental Setting

For both generator and discriminator, we employed the Adam optimizer [34] with learning rate = 0.0002. For each oversampling technique, we generated 1500 minority class samples. We trained the GAN for 3000 epochs, using Tensorflow [35] framework for deep learning and python for the classifiers. We used Sklearn [36] library to employ all the machine learning classifiers. For the purpose of implementation OCSVM [4], we used Sklearn [36] library. For different classifiers, various parameters are taken into account. In RF we choose the most critical parameters, such as no. of estimators (no of trees used to make a decision) = 100, criterion = gini, max_features = log2, minimum number of samples for leaf = 10, maximum depth of tree = 10, minimum split size = 10. bFor MLP, we used 2 hidden layers with 16 neurons each with ReLU [39] activation function. For output layer we used sigmoid activation function and Adam [37] as optimizer with learning rate = 0.001. Grid search is used to fine tune all the hyper-parameters.

## 3.6. Evaluation Metrics

Sensitivity, specificity, accuracy, and AUC Score are the metrics used to assess the success of our suggested techniques.

$$\text{Sensitivity} = TP / (TP + FN)$$

$$\text{Specificity} = TN / (TN + FP)$$

$$\text{Accuracy} = (TP + TN) / (TP + TN + FP + FN)$$

$$\text{AUC} = (\text{Sensitivity} + \text{Specificity}) / 2$$

Where, True Positive (TP): Predicted positive and actual is also positive; True Negative (TN): Predicted negative and actual is also negative; False Positive (FP): Predicted positive but actual is negative; False Negative (FN): Predicted negative but actual is positive.

Here, we ranked the classifier according to the score of AUC, since AUC is the comprehensive measure of the ability of the classifier to distinguish between the classes. Higher the AUC, the better the performance of the model in distinguishing between positive and negative classes.

## 4. Dataset Description

### 4.1. Churn prediction in bank credit card customers

For the churn prediction for bank credit card users, data is taken from a Latin American bank that was seeing a spike in customers churn and sought to improve its retention strategy. This data comes from the Business Intelligence Cup (2004). This dataset consists of 14814 instances of which 13812 are loyal clients i.e., 93%, and 1002 are churners, i.e., 7%. As a result, the percentage of churner's vs non-churners in the sample is significantly disproportionate. The complete attribute information is provided in Table 4.

**Table 1** Train and Test data Of Churn Prediction Dataset

| Data | Churner | Non-Churner | Total |
|---|---|---|---|
| Train Set | 802 | 11049 | 11851 |
| Test Set | 200 | 2763 | 2963 |

### 4.2. Insurance Fraud Detection

Pyle (1999)'s dataset is also heavily unbalanced, with 94 % legal and 6 % fradulent consumers. The only publicly available fraud detection dataset in the vehicle insurance market is provided by Angoss Knowledge Seeker software. It was originally known as "carclaims.txt" and may be found on Pyle's accompanying compact CD (1999). Collected between January 1994 to December 1995, there are 11,338 records, and from January 1996 to December 1996, there are 4,083 records. With an average of 430 claims each month, it has 6% fraudulent and 94 % legal cases. There are six numerical and 25 categorical features in the original dataset (fraud or legal).

**Table 2** Train and Test data of Insurance Fraud Dataset

| Data | Fraud | Legal | Total |
|---|---|---|---|
| Train Set | 738 | 1598 | 12336 |
| Test Set | 185 | 2899 | 3084 |

## 4.3. Loan Default Dataset

The data is related with direct marketing campaigns of a Portuguese banking institution. The classification goal is to predict if the client will subscribe a term deposit. The dataset contains 45,211 records. It has 88% of negative records and 12% of positive records. The original dataset has 17 attributes out of which 6 numerical attributes and 9 categorical attributes, including the binary class label (yes or no).

**Table 3** Train and Test data of Loan Default Dataset

| Data | Yes | No | Total |
|---|---|---|---|
| Train Set | 4231 | 31937 | 36168 |
| Test Set | 1058 | 7985 | 9043 |

**Table 4** Attribute information of churn prediction dataset

| S.No | Attribute Name | Description |
|---|---|---|
| 1 | Target | Target variable (Churn or Non-Churn) |
| 2 | CRED_T | Credit in month T |
| 3 | CRED_T-1 | Credit in month T-1 |
| 4 | CRED_T-2 | Credit in month T-2 |
| 5 | NCC_T | Number of credit cards in month T |
| 6 | NCC_T-1 | Number of credit cards in month T-1 |
| 7 | NCC_T-2 | Number of credit cards in month T-2 |
| 8 | INCOME | Customer's Income |
| 9 | N_EDUC | Customer's education level |
| 10 | Age | Customer's age |
| 11 | Sex | Customer's sex |
| 12 | E_CIV | Civilian status |
| 13 | T_WEB_T | Number of web truncation in month T |
| 14 | T_WEB_T-1 | Number of web truncation in month T-1 |
| 15 | T_WEB_T-2 | Number of web truncation in month T-2 |
| 16 | MAR_T | Customer's margin for the company in month T |
| 17 | MAR_T-1 | Customer's margin for the company in month T-1 |
| 18 | MAR_T-2 | Customer's margin for the company in month T-2 |
| 19 | MAR_T-3 | Customer's margin for the company in month T-3 |
| 20 | MAR_T-4 | Customer's margin for the company in month T-4 |
| 21 | MAR_T-5 | Customer's margin for the company in month T-5 |
| 22 | MAR_T-6 | Customer's margin for the company in month T-6 |

**Table 5** Attribute information of Pre-processed Insurance fraud dataset

| S.No | Attribute Name | Description |
|---|---|---|
| 1 | Gap | Time difference of accident and insurance claim |
| 2 | Make | Manufacturer of car |
| 3 | Accident area | Rural or urban |
| 4 | Gender | Male of female |
| 5 | Marital status | Single, married, widowed and divorced |
| 6 | Fault | Policy holder or third party |
| 7 | Policy type | Type of policy (1-9) |
| 8 | Vehicle category | Sedan, port or utility |
| 9 | Vehicle price | Price of vehicle |
| 10 | Rep. number | ID of the person who process the claim |
| 11 | Deductible | Amount to be deducted before claim disbursement |
| 12 | Driver rating | Driving experience |
| 13 | Days: policy accident | Days left in policy when accident happened |
| 14 | Days: policy claim | Days left in policy when claim was filed |
| 15 | Past number of claims | Past number of claims |
| 16 | Age of vehicle | Vehicle's age with 8 categories |
| 17 | Age of policy holder | Policy holder's age |
| 18 | Policy report filed | Yes or no |
| 19 | Witness presented | Yes or no |
| 20 | Agent type | Internal or external |
| 21 | Number of supplements | Number of supplements |
| 22 | Address change claim | No of times change of address requested |
| 23 | Number of cars | Number of cars |
| 24 | Base policy | All perils, collision or liability |
| 25 | Class | Fraud found (yes or no) |

**Table 6** Attribute information of Loan default dataset

| S.No | Attribute Name | Description |
|---|---|---|
| 1 | Age | Customer's Age |
| 2 | Job | Type of Job 12 categories |
| 3 | Marital | Divorced, Married, Single, and married |
| 4 | Education | Types of education 8 categories |
| 5 | Defaulted | Yes, no and unknown |
| 6 | Balance | Balance in account |
| 7 | Housing | Yes or no |
| 8 | Loan | Yes or no |
| 9 | Contact | Cellular or telephone |
| 10 | Day | Last contact day of the month |
| 11 | Month | Last contact month of the year |
| 12 | Duration | Last contact duration |
| 13 | Campaign | Number of contacts performed during this campaign |
| 14 | Pdays | Number of days passed by which client was last contacted |
| 15 | Previous | Number of contacts performed before this campaign and for this client |
| 16 | Poutcome | Outcome of previous campaign |
| 17 | Target | Yes or no |

## 5. Results and Discussion

We evaluated our proposed model CTGAN, OCSVM+GAN using validation data, presented in the Tables 7 through 12. We used 10-fold cross-validation, and found that our suggested oversampling technique outperforms nearly all prior work. One can notice the difference in AUC score on churn prediction dataset of our result and the previous work done by Farquad et al [5], Sundarkumar and Ravi [31], Sundarkumar, Ravi, and Siddheshwar [6] compared in Table 7 and Table 8. A similar trend is observed when we proposed a hybrid approach to balance the unbalanced data using GAN + OCSVM . We obtained high AUC scores compared to previous work done on Churn Prediction and Insurance Fraud datasets. In our experiments, we discovered that CTGAN outperforms other GANs by a significant margin, demonstrating that CTAGN is considerably better at producing structured synthetic samples than vanilla GAN and WGAN. One of the main reasons for CTAGN's superiority is that it can capture multimodal representations in continuous variables. But, either the vanilla GAN or WGAN cannot account for all the modes of continuous variables. CTAGN proposes a mode-specific normalization approach for dealing with the problem of multiple modes in continuous variables. CTGAN also deals with unbalanced data in categorical variables. To address this issue, it employs training by sampling and conditional generators, ensuring that all discrete variable categories have a reasonable chance of being included in the sample from which GAN learns. The linear and non-linear correlation between the variables is also preserved by CTAGN [3].

We examine the performance of our propsoed method in terms of AUC score, which is a suitable measure for evaluating model performance when both sensitivity and specificity are given equal weight. In Table 7, it is observed that DT, Random Forest, xgboost and LGBM yielded AUC of 0.872, 0.872, 0.862 and 0.877 respectively using CTGAN [17]. However, when we compare the results of other two classifiers LR and SVM with the results of Sundarkumar, Ravi, and Siddheshwar [6], Farquad et al. [5] there is no such significant improvement in the AUC score. One reason for the high AUC score when using tree-based classifiers is because of ensemble techniques used in Random Forest and xgboost. When we employed the Vanilla GAN (see Table 8) and the WGAN (see Table 8) for oversampling, we saw almost the same pattern. Across all datasets, t-test with (10+10-2=18) 18 degrees of freedom and a 1% level of significance is performed on the AUC to compare the classifiers. At a 1% level of significance, the table value of the t-statistic for 18 degrees of freedom is 2.83. That is, if the calculated t-statistic is more than 2.83, we may conclude that the difference between techniques is statistically significant and otherwise not. We observed that all ensemble-based models are statistically equivalent in terms of AUC as per the t-test.

In addition, we also worked on the auto insurance fraud detection data. We show the outcomes of the proposed approach in Table 9 and Table 10. In Table 9 it is observed that DT, RF, xgboost, LGBM yielded AUC of 0.739, 0.762, 0.768 and 0.76 respectively using CTGAN. When we compare the results of other two classifiers LR and SVM with the results of Sundarkumar, Ravi and Siddheshwar [6], Farquad et al.[5] we got the AUC score of 0.76 and 0.765 respectively. According to the t-test on AUC, xgboost, we discovered that there is no statistically significant difference between xgboost & Random Forest. Table 10 includes the results of additional GAN variations such as Vanilla GAN and WGAN.

On the bank default prediction dataset, Tables 11 and 12 present the results obtained. In this dataset also, the CTGAN outperformed the other GAN variants vanilla GAN and WGAN. In Table 11 it observed that CTGAN with xgboost yielded the best AUC of 0.874.

In the case of churn prediction dataset, rules obtained by DT with an AUC of 87.2 with just 9 rules are presented in Table 13. The significant outcome of the study is we obtained less number of rules with respect to [5, 6]. Sundarkumar and Ravi [31] by applying DT achieved AUC of 0.852 with 13 rules. Vasu and Ravi [37] by applying DT achieved 0.854 AUC with 14 rules. Farquad et al. [28] by applying SVM + NBTree achieved 0.794 AUC with ten rules. We recommend our proposed approach as it yields higher AUC with less number of rules compared to previous works.

In the case of the Insurance fraud detection dataset, we obtained the AUC of 74 with 5 rules using DT (see Table 14). Vasu & Ravi [37] obtained 0.711 AUC with 19 rules by using DT. Farquad et al. [28] using DT, obtained 0.711 using 4 rules. However, Sundarkumar and Ravi [31], obtained 4 rules. Sundarkumar & Ravi & Siddeshwar [6] obtained 7 rules. In Table 15, we presented the rules obtained by DT in the case of the Loan Default dataset, with an AUC of 0.849 achieved by 9 rules.

**Table 7.** Average Result on Churn Prediction Validation Dataset

| | Sundarkumar & Ravi [31] | Sundarkumar, Ravi & Siddheshwar [6] | | Farquad et al.[5] | | CTGAN | | | | CTGAN + OCSVM (Sigmoid Kernel) | | | |
|---|---|---|---|---|---|---|---|---|---|---|---|---|---|
| Classifier | AUC | AUC | Classifier | AUC | Classifiers | Spec* | Sen* | AUC | T-test | Spec* | Sen* | AUC | T-test |
| LR | 0.839 | 0.837 | ALBA | 0.7825 | LR | 0.923 | 0.77 | 0.85 | 9.68 | 0.90 | 0.785 | 0.834 | 13 |
| DT | 0.80 | 0.8425 | ALBA (SVs) | 0.7926 | DT | 0.90 | 0.83 | **0.872** | 3.0 | 0.89 | 0.84 | **0.867** | 5.3 |
| MLP | 0.614 | 0.779 | MALBA | 0.794 | MLP | .937 | 0.77 | **.853** | 6.8 | 0.90 | 0.81 | **0.855** | 5.4 |
| SVM | 0.809 | 0.872 | SVM | 0.785 | SVM | 0.92 | 0.75 | 0.83 | 15.5 | 0.89 | 0.79 | 0.842 | 18 |
| PNN | 0.6496 | 0.6113 | MALBA (normal) | 0.7933 | Xgboost | 0.975 | 0.75 | **0.86** | 4.9 | 0.96 | 0.77 | **0.866** | 6.2 |
| GMDH | 0.8477 | 0.848 | MALBA (logistic) | 0.792 | LGBM | 0.94 | 0.82 | **0.877** | | 0.93 | 0.82 | **0.878** | |
| | | | | | RF | 0.934 | 0.81 | **0.8722** | 0.4 | 0.874 | 0.88 | **0.877** | 2.8 |

Spec* = Specificity, Sen* = Sensitivity, AUC = Area under ROC Curve

**Table 8.** Average Result on Churn Prediction Validation Dataset

| | Vanilla GAN | | | | GAN+OCSVM(Sigmoid Kernel) | | | | WGAN | | | | WGAN+OCSVM(Sigmoid Kernel) | | | |
|---|---|---|---|---|---|---|---|---|---|---|---|---|---|---|---|---|
| Classifier | Spec* | Sen* | AUC | T-test | Spec* | Sen* | AUC | T-test | Spec* | Sen* | AUC | T-test | Spec* | Sen* | AUC | T-test |
| Random Forest | 0.95 | 0.78 | **0.86** | 0.72 | 0.8835 | 0.85 | **0.866** | | 0.946 | 0.80 | **0.873** | | 0.90 | 0.84 | **0.871** | 3.38 |
| Xgboost | 0.98 | 0.715 | 0.849 | 8.01 | 0.966 | 0.74 | **0.852** | 8.3 | 0.98 | 0.705 | 0.844 | 16.8 | 0.98 | 0.74 | **0.855** | 4.22 |
| Decision Tree | 0.926 | 0.785 | **0.855** | 5.8 | 0.899 | 0.74 | **0.857** | 5.1 | 0.92 | 0.785 | **0.853** | 7.71 | 0.927 | 0.81 | **0.868** | 1.3 |
| LGBM | 0.96 | 0.79 | **0.876** | | 0.80 | 0.88 | 0.841 | 5.77 | 0.973 | 0.77 | **0.87** | 2.38 | 0.94 | 0.805 | **0.873** | |
| LR | 0.86 | 0.775 | 0.818 | 19.1 | 0.846 | 0.78 | 0.813 | 30.2 | 0.8936 | 0.78 | 0.836 | 28.7 | 0.88 | 0.78 | 0.833 | 10.6 |
| SVM | 0.88 | 0.785 | 0.833 | 14.1 | 0.869 | 0.785 | 0.829 | 27.2 | 0.8693 | 0.785 | 0.828 | 37.4 | 0.8664 | 0.785 | 0.825 | 13.3 |
| MLP | 0.94 | 0.67 | 0.80 | 12.4 | 0.94 | 0.745 | 0.842 | 6.4 | 0.948 | 0.72 | 0.834 | 12.4 | 0.938 | 0.74 | 0.84 | 7.8 |

Spec* = Specificity, Sen* = Sensitivity, AUC = Area under ROC Curve

**Table 9.** Average Result on Insurance Fraud Validation Dataset

| | Sundarkumar and Ravi [31] | Sundarkumar, Ravi and Siddheshwar [6] | Farquad et al.[5] | | CTGAN | | | | CTGAN + OCSVM(Sigmoid Kernel) | | | |
|---|---|---|---|---|---|---|---|---|---|---|---|---|
| Classifier | AUC | AUC | Classifier | AUC | Classifiers | Spec* | Sen* | AUC | T-test | Spec* | Sen* | AUC | T-test |
| LR | 0.745 | 0.718 | ALBA | 0.7213 | LR | | 0.552 | **0.76** | 22.7 | 0.549 | 0.97 | 0.75 | 30 |
| DT | 0.7322 | 0.76 | ALBA (SVs) | 0.7178 | DT | 0.58 | 0.89 | 0.74 | 18.3 | 0.56 | 0.945 | 0.756 | 10 |
| MLP | 0.68 | 0.6374 | MALBA | 0.7193 | MLP | 0.67 | 0.77 | 0.72 | 12.4 | 0.75 | 0.643 | 0.70 | 9.7 |
| SVM | 0.747 | 0.751 | SVM | 0.720 | SVM | 0.962 | 0.566 | **0.765** | 10.4 | 0.56 | 0.96 | 0.77 | 11 |
| PNN | 0.6853 | 0.735 | MALBA (normal) | 0.718 | Xgboost | 0.574 | 0.96 | **0.768** | | 0.593 | 0.95 | **0.77** | 1.4 |
| GMDH | 0.7514 | 0.725 | MALBA (logistic) | 0.7146 | LGBM | 0.615 | 0.924 | **0.76** | 18.4 | 0.60 | 0.92 | 0.762 | 13 |
| | | | | | RF | 0.576 | 0.962 | **0.762** | 3.2 | 0.584 | 0.96 | **0.773** | |

Spec* = Specificity, Sen* = Sensitivity, AUC = Area under ROC Curve

**Table 10.** Average Result ON Insurance Fraud Validation Dataset

| | Vanilla GAN | | | | GAN+OCSVM (Sigmoid Kernel) | | | | WGAN | | | | WGAN+OCSVM (Sigmoid Kernel) | | | |
|---|---|---|---|---|---|---|---|---|---|---|---|---|---|---|---|---|
| Classifier | Spec* | Sen* | AUC | T-test | Spec* | Sen* | AUC | T-test | Spec* | Sen* | AUC | T-test | Spec* | Sen* | AUC | T-test |
| Random Forest | 0.605 | 0.929 | **0.76** | 3.4 | 0.613 | 0.913 | 0.763 | 0.6 | 0.40 | 0.848 | 0.62 | 25.8 | 0.61 | 0.91 | 0.762 | 1.45 |
| Xgboost | 0.84 | 0.475 | 0.65 | 12.63 | 0.60 | 0.924 | 0.765 | 4.53 | 0.28 | 0.97 | 0.63 | 20.1 | 0.611 | 0.918 | 0.765 | 3.85 |
| Decision Tree | 0.67 | 0.82 | **0.751** | 1.13 | 0.616 | 0.90 | **0.759** | 1.1 | 0.375 | 0.767 | 0.571 | 24 | 0.61 | 0.90 | 0.759 | 5.8 |
| LGBM | 0.70 | 0.778 | 0.74 | 1.3 | 0.629 | 0.864 | 0.747 | 5.85 | 0.44 | 0.69 | 0.57 | 21.3 | 0.636 | 0.886 | 0.761 | 5.77 |
| LR | 0.9621 | 0.566 | **0.76** | | 0.567 | 0.962 | **0.764** | | 0.96 | 0.5674 | **0.764** | | 0.5685 | 0.96 | **0.767** | |
| SVM | 0.8378 | 0.654 | 0.749 | 8.4 | 0.6109 | 0.891 | 0.755 | 9.39 | 0.83 | 0.65 | 0.74 | 11.28 | 0.60 | 0.87 | 0.755 | 8.1 |
| MLP | 0.721 | 0.729 | 0.725 | 7.47 | 0.85 | 0.43 | 0.64 | 14.6 | 0.74 | 0.664 | 0.70 | 12.4 | 0.84 | 0.46 | 0.65 | 9.7 |

Spec* = Specificity, Sen* = Sensitivity, AUC = Area under ROC Curve

**Table 11.** Average Result on Loan Default Validation Dataset

| | CTGAN | | | | CTGAN+OCSVM (Sigmoid Kernel) | | | | Vanilla GAN | | | | GAN+OCSVM (Sigmoid Kernel) | | | |
|---|---|---|---|---|---|---|---|---|---|---|---|---|---|---|---|---|
| Classifier | Spec* | Sen* | AUC | T-test | Spec* | Sen* | AUC | T-test | Spec* | Sen* | AUC | T-test | Spec* | Sen* | AUC | T-test |
| Random Forest | 0.846 | 0.85 | **0.849** | 14.1 | 0.846 | 0.859 | **0.852** | 6.25 | 0.824 | 0.845 | **0.835** | | 0.855 | 0.862 | **0.859** | |
| Xgboost | 0.91 | 0.76 | 0.837 | 12.4 | 0.90 | 0.759 | 0.834 | 17.6 | 0.82 | 0.744 | 0.783 | 20.1 | 0.918 | 0.75 | 0.83 | 15.1 |
| Decision Tree | 0.82 | 0.851 | 0.836 | 14.5 | 0.80 | 0.879 | **0.843** | 12.2 | 0.7562 | 0.80 | 0.78 | 6.5 | 0.835 | 0.853 | 0.844 | 10.5 |
| LGBM | 0.824 | 0.92 | **0.874** | | 0.90 | 0.82 | **0.86** | | 0.804 | 0.804 | 0.804 | 12.4 | 0.906 | 0.80 | **0.857** | 0.01 |
| LR | 0.776 | 0.857 | 0.81 | 28.3 | 0.777 | 0.85 | 0.81 | 32.4 | 0.755 | 0.879 | 0.817 | 19.4 | 0.7606 | 0.8733 | 0.8169 | 35.6 |
| SVM | 0.75 | 0.892 | 0.825 | 31.4 | 0.7626 | 0.88 | 0.82 | 30.1 | 0.765 | 0.888 | 0.826 | 7.23 | 0.766 | 0.887 | 0.82 | 32.3 |
| MLP | 0.78 | 0.87 | 0.831 | 12.3 | 0.786 | 0.876 | 0.83 | 16.1 | 0.85 | 0.792 | 0.821 | 7.3 | 0.81 | 0.85 | 0.83 | 5.8 |

Spec* = Specificity, Sen* = Sensitivity, AUC = Area under ROC Curve

**Table 12.** Average Result on Loan Default Validation Dataset

| | WGAN | | | | WGAN+OCSVM (Sigmoid Kernel) | | | |
|---|---|---|---|---|---|---|---|---|
| Classifiers | Spec* | Sen* | AUC | T-test | Spec* | Sen* | AUC | T-test |
| Random Forest | 0.83 | 0.836 | **0.833** | | 0.8647 | 0.85 | **0.859** | |
| Xgboost | 0.814 | 0.73 | 0.774 | 16.4 | 0.92 | 0.728 | 0.82 | 24.9 |
| Decision Tree | 0.743 | 0.812 | 0.778 | 7.8 | 0.83 | 0.85 | 0.84 | 8.36 |
| LGBM | 0.806 | 0.792 | 0.799 | 10.54 | 0.9078 | 0.80 | **0.85** | 0.46 |
| LR | 0.742 | 0.872 | 0.807 | 20.71 | 0.7381 | 0.871 | 0.80 | 40.3 |
| SVM | 0.7523 | 0.889 | 0.82 | 8.1 | 0.752 | 0.886 | 0.819 | 29.86 |
| MLP | 0.82 | 0.80 | 0.81 | 5.1 | 0.84 | 0.82 | 0.83 | 7.91 |

Spec* = Specificity, Sen* = Sensitivity, AUC = Area under ROC Curve

**Table 13.** Rules Obtained by Decision Tree on Churn Prediction Dataset

| Rule # | Antecedents | Consequents |
|---|---|---|
| 1 | If (CRED_T<=595.23) and (NCC_T<=0.50) and (MAR_T-3<=11.36) | Churn |
| 2 | If(CRED_T>592.76) and (MAR_T<=0.04) | Churn |
| 3 | If (CRED_T<=595.23) and (NCC_T>0.50) and (MAR_T<=7.87) | Churn |
| 4 | If (CRED_T<=595.23) and (NCC_T>0.50) and (MAR_T > 7.87) | Non-Churner |
| 5 | If(CRED_T > 592.76) and (MAR_T-6 > -3.25) | Non-Churner |
| 6 | If (CRED_T > 595.23) and (NCC_T<=0.50) and (MAR_T <= -5.23) and (MAR_T-4 <= 5.50) | Non-Churner |
| 7 | If (CRED_T > 595.23) and (NCC_T<=0.50) and (MAR_T <= -5.23) and (MAR_T-4 > 5.50) | Churner |
| 8 | If (CRED_T > 595.23) and (NCC_T > 0.50) and (CRED_ <= 598.59) | Non-Churner |
| 9 | If (CRED_T > 595.23) and (NCC_T > 0.50) and (MAR_T-3 <= 18.53) | Non-churner |

**Table 14.** Rules Obtained by Decision Tree on Insurance Fraud Dataset

| Rule # | Antecedents | Consequents |
|---|---|---|
| 1 | If (BasePolicy <= 1.50) and (Vehicle Category <= 1.50) and (PolicyType <= 4.50) | Fraud |
| 2 | If (BasePolicy <= 1.50) and (Vehicle Category <= 1.50) and (PolicyType > 4.50) | Non-Fraud |
| 3 | If (BasePolicy > 1.50) and (Fault <= 0.50) and (PolicyType <= 5.50) | Fraud |
| 4 | If (BasePolicy > 1.50) and (Fault > 0.50) and (AddressChange-Claim <= 1.50) | Non-Fraud |
| 5 | If (BasePolicy > 1.50) and (Fault > 0.50) and (AddressChange-Claim > 1.50) | Fraud |

**Table 15.** Rules Obtained by Decision Tree on Loan Default Dataset

| Rule # | Antecedents | Consequents |
|---|---|---|
| 1 | If (duration <= 205.50) and (campaign <= 2.50) and (month < 3) | Non-Defaulter |
| 2 | If (duration <= 205.50) and (month > 9.5) | Defaulter |
| 3 | If (duration > 88.50) and (contact <= 1.50) and (poutcome <= 2.50) | Defaulter |
| 4 | If (duration > 88.50) and (contact <= 1.50) and (poutcome > 2.50) and (age <= 43.50) | Non-Defaulter |
| 5 | If (duration > 88.50) and (contact > 1.50) and (month <= 8.50) | Defaulter |
| 6 | If (duration > 88.50) and (contact > 1.50) and (month > 8.50) | Non-Defaulter |
| 7 | If (duration > 205.50) and (contact <= 1.50) and (housing <= 0.50) | Defaulter |
| 8 | If (duration > 472.50) and (balance <= 5795.50) | Defaulter |
| 9 | If (duration > 472.50) and (balance > 5795.50) | Non-Defaulter |

## 6. Conclusion and Future Work

In this paper, we proposed FinGAN, a strategy for oversampling the minority class in three unbalanced, severely skewed datasets occurring in analytical CRM in order to improve the performance of classifiers. In our paper, we show that GAN and GAN variants based oversampling may help a classifier perform better.

      We also demonstrated a hybrid approach of oversampling and undersampling of minority and majority classes respectively. There, first we oversampled the minority class data using GAN and then extracted the support vectors from majority class by using OCSVM. We compared the efficacy of our technique with several classifiers such as LR, SVM, RF, DT, LGBM, and xgboost on a unbalanced validation dataset that is similar to the real-world situation. The efficacy of our technique was demonstrated on churn prediction, auto insurance fraud detection and loan default prediction datasets. Results showed that CTGAN based approaches yielded best statistically significant results. In future, we can use some other variant of GAN with the combination of some other powerful classifiers such as TabNet [38] in order to achieve better AUC scores.

15. M.J. Kusner and J.M. Hernández-Lobato. Gans for sequences of discrete elements with the gumbel-softmax distribution. https://arxiv.org/abs/1611.04051
16. H. Ping, J. Stoyanovich, and B. Howe. Data synthesizer: Privacy-preserving synthetic datasets. In Proceedings of the 29th International Conference on Scientific and Statistical Database Management, page 42. ACM, 2017.
17. C. Esteban, S.L Hyland, and G. Rätsch. Real-valued (medical) time series generation with recurrent conditional Gans. https://arxiv.org/abs/1706.02633
18. R. Camino, C. Hammer-schmidt, R. State. Generating multi-categorical samples with generative adversarial networks. https://arxiv.org/abs/1807.01202
19. E. Choi, S. Biswal, B. Malin, J. Duke., W.F. Stewart, and J. Sun, 2017. Generating multi-label discrete patient records using generative adversarial networks. https://arxiv.org/abs/1703.06490
20. S. Patel, A. Kakadiya, M. Mehta, R. Derasari, R. Patel, and R. Gandhi, Correlated discrete data generation using adversarial training. https://arxiv.org/abs/1804.00925
21. N. Park, M. Mohammadi, K. Gorde, S. Jajodia, H. Park, and Y. Kim. Data synthesis based on generative adversarial networks. Proceedings of the VLDB Endowment, 11(10):1071–1083, 2018.
22. L. Xu, K. Veeramachaneni, Synthesizing Tabular Data using Generative Adversarial Networks. https://arxiv.org/pdf/1811.11264
23. K.A. Smith, J.N.D. Gupta, Neural networks in business: techniques and applications for the operations researcher 27 (2000) 1023-1044.
24. J.B. Ferreira, M. Vellasco, M.A. Pacheco, C.H Barbosa, 2Oß4. Data mining techniques on the evaluation of wireless churn. In: (ESANN'2004). Proceedings European Symposium on Artificial Neural Networks Bruges (Belgium), d-sidepublication ISBN 2-930307-04-8, pp. 483-488.
25. D.C. Kumar, V. Ravi, 2008, Predicting credit card customer churn in banks using data mining. Int. J. Data Anal. Tech. Strat. 1 (1), 4-28.
26. N.V. Chawla, K.W. Bowyer, L.O. Hall, W.P. Kegelmeyer, SMOTE Synthetic Minority Over-sampling Technique, Journal of Artificial Intelligence Research 16 (2002) 321–357.
27. B. Larivie're, D.V. den Poel, investigating the role of product features in preventing customer churn, by using survival analysis and choice modelling: The case of financial services. Expert Syst. Application, 27 (2). 277-285.
28. O.G. Ali, and U. ArÕtürk, "Dynamic churn prediction framework with more effective use of rare event data: The case of private banking", Expert Systems with Applications, Vol.41 (17), pp. 7880-7903, 2014.
29. W. Verbeke, D. Martens, C. Mues, and B. Baesens, "Building comprehensible customer churn prediction models with advanced rule induction techniques", Expert Systems with Applications, Vol. 38 (3), pp.2354–2364, 2011.
30. C. F. Tsai, and Y. H. Lu, "Customer churn prediction by hybrid neural networks", Expert Systems with Applications, Vol. 36 (10), pp. 12547- 12553, 2009.